\documentclass[conference]{IEEEtran}
\IEEEoverridecommandlockouts
\usepackage{cite}
\usepackage{amsmath,amssymb,amsfonts}
\usepackage{algorithmic}
\usepackage{graphicx}
\usepackage{textcomp}
\usepackage{xcolor}
\usepackage{comment}
\usepackage{subfigure}

\def\BibTeX{{\rm B\kern-.05em{\sc i\kern-.025em b}\kern-.08em
    T\kern-.1667em\lower.7ex\hbox{E}\kern-.125emX}}
\begin{document}

\title{Segment Anything Meets Semantic Communication}

\author{
\IEEEauthorblockN{
Shehbaz Tariq,
Brian Estadimas Arfeto,
Chaoning Zhang,
Hyundong Shin, \IEEEmembership{Fellow,~IEEE}
}

\thanks{
This work was supported by the National Research Foundation of Korea (NRF) grant funded by the Korea government (MSIT) (NRF-2019R1A2C2007037, NRF-2022R1A4A3033401) and by the MSIT (Ministry of Science and ICT), Korea, under the ITRC (Information Technology Research Center) support program (IITP-2023-2021-0-02046) supervised by the IITP (Institute for Information \& Communications Technology Planning \& Evaluation).
}
\thanks{
S.~Tariq, B.~E.~Arfeto, C.~Zhang,
and H.~Shin 
are with the Department of Electronics and Information Convergence Engineering,
Kyung Hee University,
1732 Deogyeong-daero, Giheung-gu,
Yongin-si, Gyeonggi-do 17104 Korea
(e-mail: hshin@khu.ac.kr). 
}
}

\maketitle

\begin{abstract}
In light of the diminishing returns of traditional methods for enhancing transmission rates, the domain of semantic communication presents promising new frontiers. Focusing on image transmission, this paper explores the application of foundation models, particularly the Segment Anything Model (SAM) developed by Meta AI Research, to improve semantic communication. SAM is a promptable image segmentation model that has gained attention for its ability to perform zero-shot segmentation tasks without explicit training or domain-specific knowledge. By employing SAM's segmentation capability and lightweight neural network architecture for semantic coding, we propose a practical approach to semantic communication. We demonstrate that this approach retains critical semantic features, achieving higher image reconstruction quality and reducing communication overhead. This practical solution eliminates the resource-intensive stage of training a segmentation model and can be applied to any semantic coding architecture, paving the way for real-world applications. 
\end{abstract}


\section{Introduction}

\begin{figure*}[t!]
\centering

\includegraphics[width=0.90\textwidth]{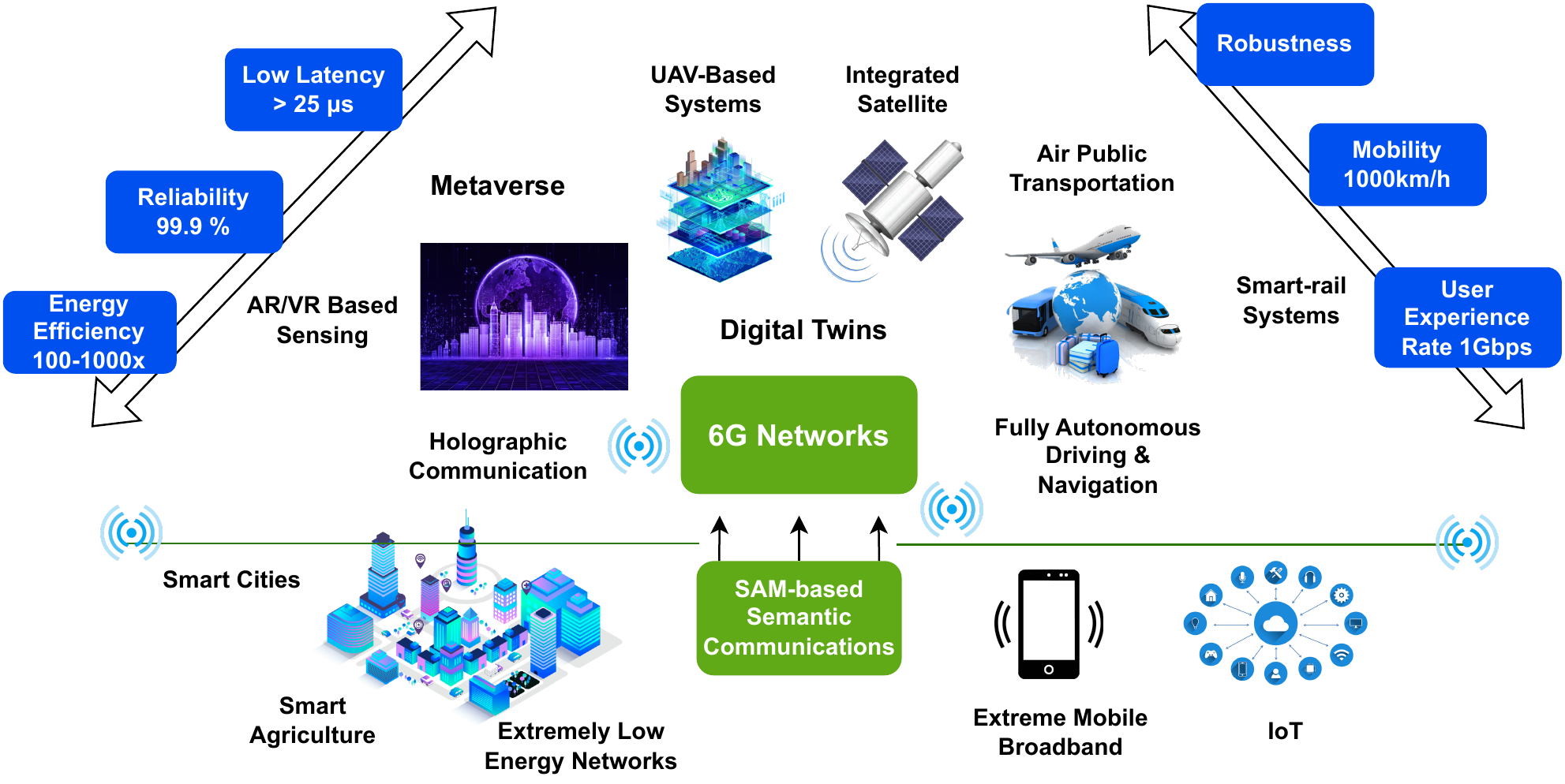}
\caption{
Depiction of the robustness, speed, and reliability underpinning 6G Networks, highlighting the potential integration of a SAM-based Semantic Communications model for diverse applications.
}
\label{fig:1}
\end{figure*}

As we approach Shannon's limits, traditional methods to enhance transmission rates, such as increasing power, expanding bandwidth, and adding antennas, are becoming less effective due to spectrum limitations and high power consumption \cite{luo2022semantic}. Consequently, exploring the semantic domain has become crucial for significantly improving communication efficiency. Semantic communication allows the compact representation of the source information at a semantic level to retain only the task-relevant information, which can then be used to recover the source data\cite{yang2023witt} or perform some intelligent tasks \cite{xie2022task}, such as classification, object detection, etc., more efficiently. The robustness of semantic communication holds significant potential for various 6G applications \cite{alwis2021survey6G}, such as the metaverse, mobile broadband, digital twins, and fully autonomous driving, as shown in Fig \ref{fig:1}. By ensuring reliability and efficiency, semantic communication can contribute to building a strong foundation for 6G systems\cite{calvan2021networks}. These systems will support seamless interactions and data exchange in the metaverse, optimize content delivery in mobile broadband, enable real-time data exchange in digital twins, and facilitate precise control in fully autonomous driving. Consequently, numerous studies have shown that deep learning-based semantic communication outperforms traditional techniques\cite{xie2021deep, huang2022toward, lan2021semantic}. 

However, a challenge persists: many current deep learning techniques require a shared knowledge base, which makes them highly task-specific and restricts their ability to generalize to downstream tasks. Building upon this, the remarkable rise of foundational models \cite{bommasani2021opportunities, amatriain2023transformer} (e.g., BERT and GPT) has been compared to seminal discoveries like electricity and the internet for humanity. These models, subjected to self-supervised learning~\cite{zhang2022survey} on a vast array of multi-modal data, demonstrate the capacity to generalize downstream tasks without explicit training using prompt engineering \cite{strobelt2022interactive}. The foundation models also significantly contribute to the development of generative AI (AIGC)~\cite{zhang2023complete}, like ChatGPT~\cite{zhang2023one} and text-to-image~\cite{zhang2023text}, text-to-3D~\cite{li2023generative}. Inspired by these successes, we explore the application of the Segment Anything Model (SAM) \cite{kirillov2023segment}, a novel foundation model for computer vision developed by Meta AI Research, for semantic communication.

SAM is a promptable image segmentation model that has gained significant attention for its ability to perform zero-shot segmentation tasks without finetuning\cite{zhang2023sam_survey}. This model allows for the segmentation of any image with just a simple prompt, thereby eliminating the need for labor-intensive image labeling and domain expertise. Moreover, it also gives the flexibility to send only the necessary segments of an image, which reduces communication overhead and enhances intelligent task achievability, such as object detection and image classification across various domains \cite{zhou2023dsec, wang2023scaling}.  Building on the remarkable generalization capability of SAM, we propose a new approach to semantic communication for image transmission that requires image segmentation, followed by a lightweight neural network architecture for joint source and channel encoding and decoding that significantly reduces the communication overhead and retains higher image quality. By taking advantage of SAM's segmentation, we demonstrate that our approach allows for preserving meaningful semantic features, resulting in higher image reconstruction quality as measured by the peak signal-to-noise ratio (PSNR) compared to the non-segmentation-based approach. Moreover, by obviating the need for domain-specific expertise and training a segmentation model, which is often the most resource-intensive stage in the semantic communication pipeline, we can plug the SAM into any semantic encoder, thus providing a pragmatic solution for real-world applications, as shown in Fig 1.

\section{Related Work}

The pioneering work in semantic communication was the deep joint source-channel coding for image transmission \cite{bourtsoulatze2019deep}, which proposed directly mapping image pixel values to channel inputs. This approach demonstrated improved PSNR performance compared to conventional compression techniques such as JPEG and JPEG2000 \cite{santa2002jpeg} under noisy channel conditions. Subsequently, an alternative approach, the Wireless Image Transmission Transformer for Semantic Communications (WITT) \cite{yang2023witt}, was introduced, leveraging a different neural network architecture based on transformers \cite{vaswani2017attention}. This approach, demonstrating robustness, further enhanced the PSNR. Nevertheless, both of these methodologies perform semantic encoding and decoding of the entire image, potentially including irrelevant information.

To address this, a Semantic Segmentation-based Semantic Communication System for image transmission \cite{wu2023semantic} was proposed, which semantically segments the image to distinguish between regions of interest (ROI) and regions of non-interest (RONI) before encoding and decoding. This approach further improved transmission accuracy over existing methods. However, the segmentation process is the bottleneck, requiring expert labeling of the image into ROI and RONI, followed by training \cite{garcia2017review}. Consequently, identifying ROI and RONI for specific task achievability remains challenging. The outlined approach necessitates training two distinct semantic encoders and decoders for each ROI, often termed the foreground and for each RONI, or the background. While potentially effective for a single ROI, this methodology presents scalability concerns when confronted with images encompassing multiple ROIs. The complexity of training individual models for each distinct region might become prohibitively high, posing constraints on the applicability and efficacy of this system for more intricate image analysis.

Therefore, we propose a pragmatic architecture in this paper that leverages the zero-shot learning capability of Segment Anything (SAM) to generate masks of the image given a prompt for the ROI. Prior works have also investigated SAM under challenging scenarios like glass (transparent objects) and adversarial attack~\cite{zhang2023sam_survey,zhang2023understanding,zhang2023attack,han2023segment}. It has been found in~\cite{zhang2023understanding} SAM is more biased towards texture than shape. A complete survey on SAM is provided in~\cite{zhang2023sam_survey}. Our work is the first to combine semantic communication with SAM, which eliminates the need for expert labeling and training on new data to get ROI. As a promptable segmentation model, it also adds flexibility to generate multiple masks. We show that SAM-based architecture eliminates the need to train for segmentation and improves performance on benchmarks similar to those in \cite{wu2023semantic}, thanks to its high-quality masks.

\section{System Model}
\begin{figure*}[t!]
\centering

\includegraphics[width=1.00\textwidth]{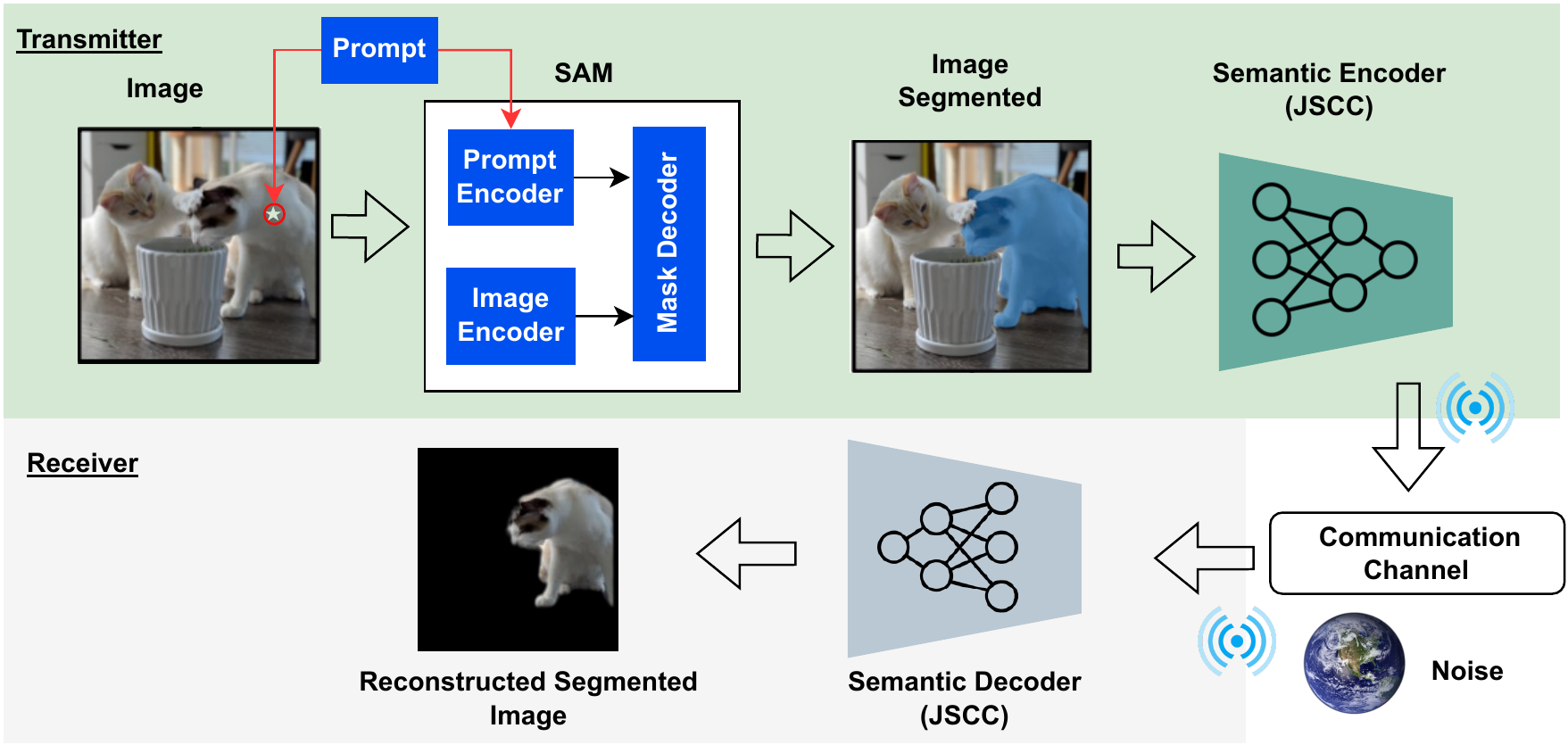}
\caption{SAM-based framework architecture for Semantic Communication. 
}
\label{fig:2}
\end{figure*}

In this section, SAM based semantic communication system is proposed for image transmission. The system model is shown in Fig \ref{fig:2}. It can be broken down into three parts. The transmitter uses SAM to segment the image and extract the semantic features of their structured representations from the generated masks and then extracts their compressed semantic features; the channel is used to simulate realistic noise; The receiver is used to reconstruct the original image from the noisy, compressed features.

\subsection{Image Transmission}
This process is separated into encoding, feature extraction, and decoding on the sender side. The result that we get is masked objects along with the background. We employ a systematic process to achieve accurate image segmentation using SAM for semantic communication. SAM mainly consists of three parts, a prompt encoder that takes in a prompt to segment a specific region and a Vision transformer(ViT) based image encoder that takes in an RGB image. The generated image embeddings are sent to a lightweight decoder with the embedded prompt.

\textbf{Image Encoding} Let $s \in \mathbb{R}^{H\times W\times C}$ be an image, where $H, W,$ and $C$ are the image height, width, and channel, respectively. The  SAM image encoder  $S_{enc}$, a trained neural network specifically designed for segmentation tasks,  transforms the $s$ into embedding $x_{e} \in \mathbb{R}^{H\times W\times C}$.

\textbf{Prompt encoding} Let $p_{i}$ be a $256$-dimensional vector that encodes the $i$th prompt, which can be a point on the image, a box around an object, a free-form text, or some shaded area over the object. SAM can also work in ``everything" mode by generating masks on almost every object on the image. 


\textbf{Mask decoder}
The embedding  $x_{e}$ of the image $s$ and the input prompt $p_i$ are then used to produce a mask $m_i$  with a lightweight decoder. These masks effectively act as semantic representations of the original image, providing a concise visual summary. By transmitting only the masks, we retain the essential information necessary for accurate image reconstruction while achieving a higher compression level, enabling more efficient data transmission.

\textbf{Joint source-channel coding (JSCC)} We employ a Convolutional Neural Network (CNN)-based architecture, comprised of an encoder $F_{enc}$ and a decoder $F_{enc}$, dedicated solely to the source-channel coding of the masks. The architecture of the model is illustrated in Fig 3. The semantic features, denoted as $x_s$, are obtained by encoding the input $x_{in}$, which could be either the input image, mask, or a batch of masks. Mathematically, this process can be expressed as:

\begin{equation}
x_s = F_{enc}(x_{in}),
\end{equation}

\begin{figure}[htbp]
\centerline{\includegraphics[width=0.45\textwidth]{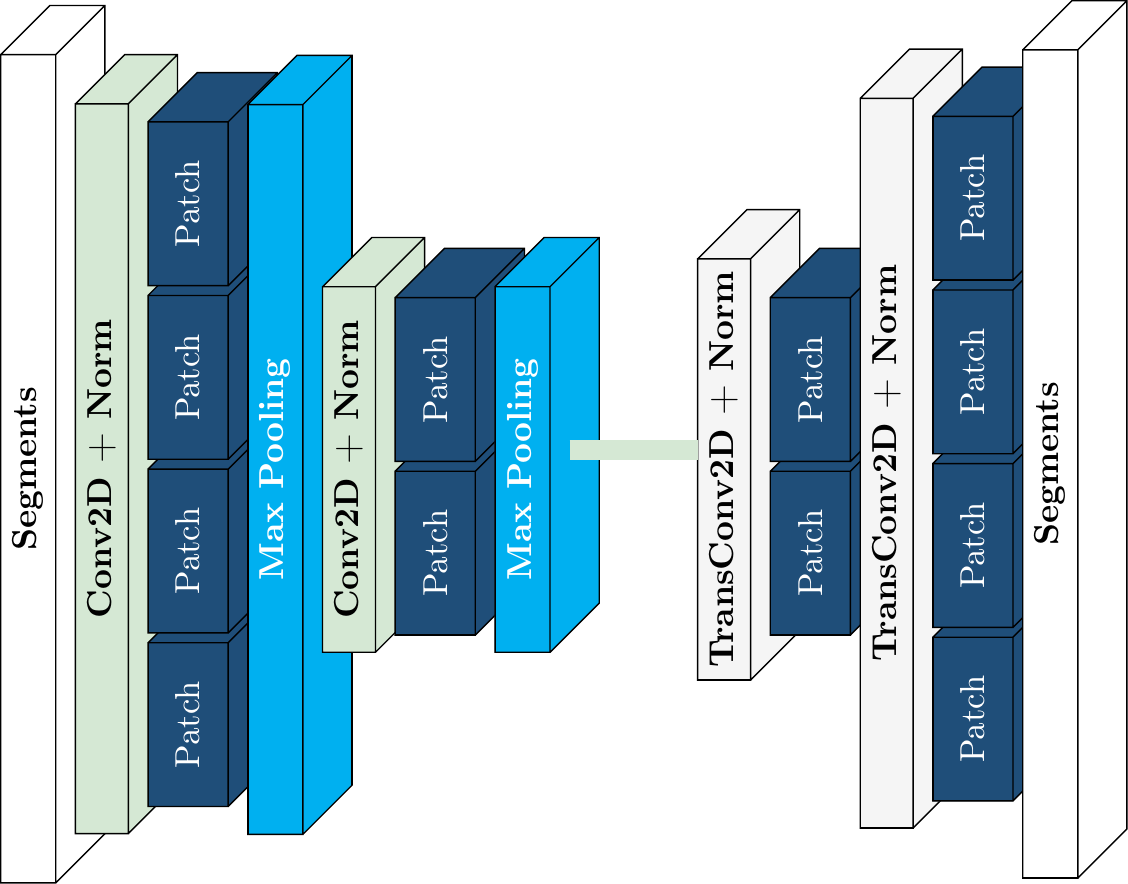}}
\caption{Semantic encoder and semantic decoder atchitectures.}
\label{fig:3}
\end{figure}

A noteworthy feature of this architecture is its ability to compress the transmitted signal size by $\mathbf{83.3\%}$, enabling more efficient communication channel utilization. The achieved compression is quantifiable via the compression ratio (CR), given by:
\begin{equation}
CR = \frac{ C_{o} \times H_{o} \times W_{o}}{C_{in} \times H_{in} \times W_{in}} ,
\end{equation}
where $C_{o}, H_{o}, W_{o}$ represent the output channels, height, and width from the semantic encoder, respectively. Conversely, $C_{in}, H_{in}, W_{in}$ denote their corresponding values for the input image.

\subsection{Channel Layer}

To establish a communication channel, we consider the presence of noise in the environment. Although the real physical noise is more complex to model for simplicity, we consider two common types of noise, additive white Gaussian noise (AWGN) and Rayleigh fading (RF), that are more widely adopted in several studies \cite{hou2001performance, bourtsoulatze2019deep, yang2023witt, wu2023semantic}. AWGN represents random noise, while RF accounts for signal attenuation due to multipath propagation.

 \begin{figure}[t]
 \centerline{\includegraphics{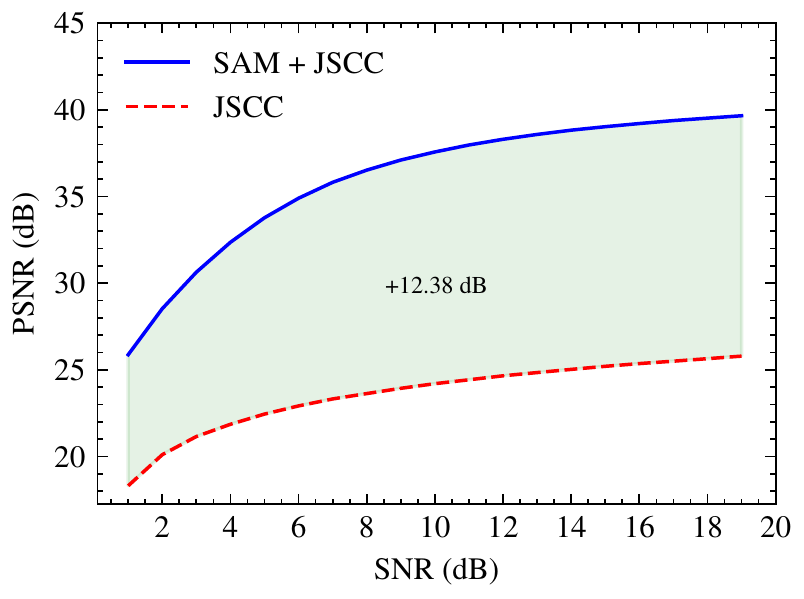}}
 \caption{Performance comparison of the image recovery of masked images and original images under AWGN noise.}
 \label{fig:4}
\end{figure}

\subsection{Image Reconstruction Block}

The noisy received signal $y \in \mathbb{R}^{H_o\times W_o\times C_o}$ is subsequently input to the CNN decoder, which reconstructs the original image, represented as $x_{in}^{\prime} \in \mathbb{R}^{H_{in}\times W_{in}\times C_{in}}$. This process is formalized as:

\begin{equation}
x_{in}^{\prime} = F_{dec}(y).
\end{equation}\label{eq:2}

This decoding operation assures accurate image reconstruction while maintaining computational efficiency.

 \begin{figure}[t]
 \centerline{\includegraphics{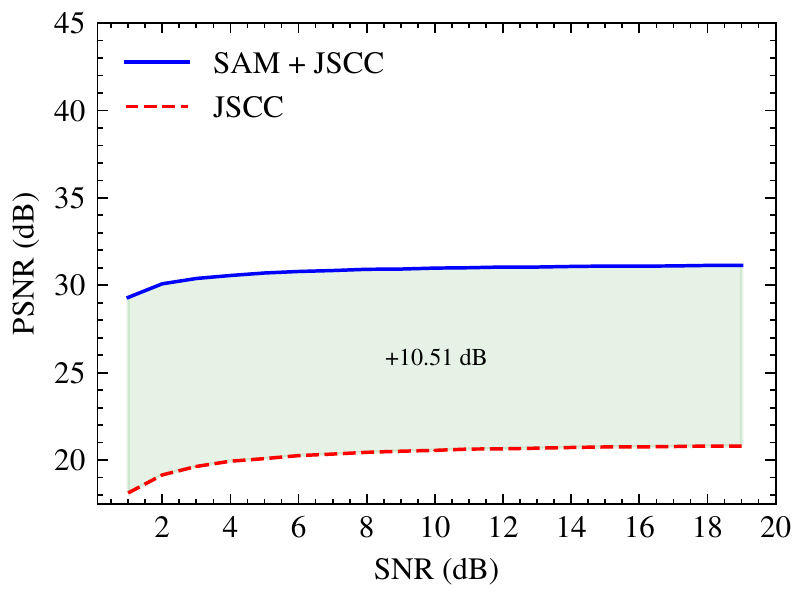}}
 \caption{Performance comparison of the image recovery of masked images and original images under RF noise.}
 \label{fig:5}
\end{figure}

\section{Results and Discussion}
In this paper, we use Pascal VOC as the dataset \cite{everingham2015pascal}. However, \textbf{SAM's zero-shot learning} advantage enables it to accurately segment and recognize objects without any training or labeled examples, which means it is ready to use for any image data. This versatility allows for efficient adaptation and effective semantic communication in real-life applications. We train on images from the dataset \textbf{without any noise} and then test them under noisy conditions. Table \ref{tab1} presents the hyperparameters used in this study.

\begin{table}[htbp]
\caption{Dataset and Hyperparameter Details}
\begin{center}
\begin{tabular}{|c|c|}
\hline
\textbf{Hyperparameters} & \textbf{Description} \\
\hline
Number of Test Images & $60$ \\
Image Resolution & $3 \times 512 \times 512$ \\
Semantic Features & $128 \times 32 \times 32$ \\
Compression Ratio & $\frac{1}{6}$ \\
Epoch & $1500$ \\
Learning Rate & $0.001$ \\
Loss Function & MSE \\
Optimizer & Adam \\
Noise Channel & AWGN - RF \\
Desired SNR & $1-20$ \\
\hline
\end{tabular}
\label{tab1}
\end{center}
\end{table}

\begin{figure}[t]
  \centering
  \begin{minipage}{.45\textwidth}
    \begin{tabular}{ccc}
      \subfigure[Original Image]
      {\includegraphics[width=.30\linewidth]{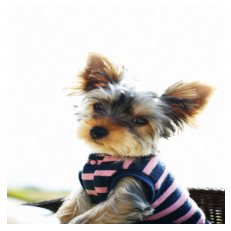}} &
      \subfigure[RF]{\includegraphics[width=.30\linewidth]{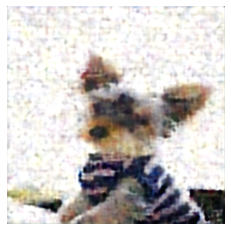}} &
      \subfigure[AWGN]{\includegraphics[width=.30\linewidth]{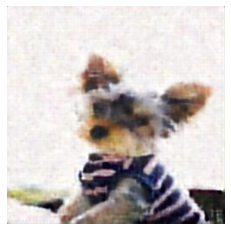}} \\
      \subfigure[ROI from SAM]{\includegraphics[width=.30\linewidth]{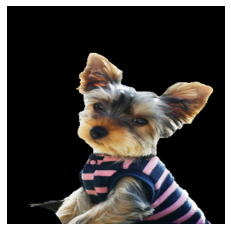}} &
      \subfigure[RF]{\includegraphics[width=.30\linewidth]{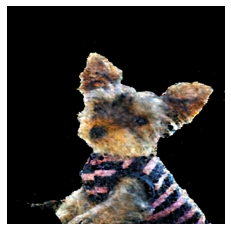}} &
      \subfigure[AWGN]{\includegraphics[width=.30\linewidth]{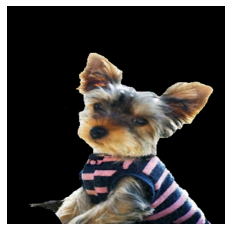}} \\
    \end{tabular}
    \caption{Visual comparison of the original image (a) and ROI (d), recovery without the use of SAM (b and c), and recovery of masks employing SAM (e and f).}
    \label{fig:6}
  \end{minipage}
\end{figure}

We train the model without using SAM on the original images, then we plug SAM and segment the images before feeding them to the network for training. We then test the image reconstruction quality in two noisy cases, AWGN and RF noise, for different SNR conditions, as shown in Fig \ref{fig:4} and fig \ref{fig:5}, using masked images for training improves the PSNR. In Fig \ref{fig:6}, we show the reconstructed images in both cases. From these results, we conclude that SAM-based segmentation helps the model to capture more features in training and thus can show more robustness to channel noise. The rationale for using SAM is its ability to segment anything without being trained. It gives us more flexibility to use this on real-life personalized datasets and only trains the semantic coding models. In this paper, for brevity, we only show the performance of the reconstruction capability of the proposed architecture. However, using SAM has even more potential applications for a practical level of communication to focus on the efficiency of task achievability, i.e., image recognition. In such specific tasks, we do not need to send the whole image but just the ROI, and SAM can achieve this in a zero-shot manner with a given prompt. The masks are more robust to noise than sending the whole image, potentially increasing task achievability.

\section{Conclusion and Future Work}
In this paper, we study the influence of segmentation on the training of semantic coding models. We presented a practical image segmentation method using SAM, eliminating the need for expert data labeling. Instead, it requires a simple prompt to segment ROI, reflecting the nature of a foundational model. Our findings show improvement in image recovery performance post-exposure to a noisy communication channel through this approach. While our focus in this paper was limited to the impact of segmentation on image recovery in semantic coding models, future research could extend this to investigate its role in task achievability. Additionally, the burgeoning potential of the transformer architecture, noted for its scalability and superior performance in computer vision tasks, could be harnessed for further investigation.

\bibliographystyle{IEEEtran}
\bibliography{IEEE-Milcom-AISC}

\end{document}